\documentclass[10pt,conference]{IEEEtran}

\usepackage[T1]{fontenc}
\usepackage[utf8]{inputenc}
\usepackage{amsmath,amssymb}
\usepackage{graphicx}
\usepackage{booktabs}
\usepackage{url}
\usepackage{cite}
\usepackage{textcomp}
\usepackage[hidelinks]{hyperref}

\begin{document}

\title{Self-organizing Architecture of Receptron Units:\\ a Hardware-Aware Framework for Edge Intelligence}

\author{
\IEEEauthorblockN{Stefano Radice, Ludovico Casaccia, Riccardo Emanuele Beccalli, Bruno Paroli, Paolo Milani}
\IEEEauthorblockA{Department of Physics ``Aldo Pontremoli'' and CIMaINa \\
University of Milano \\
Milano, Italy}
}

\maketitle

\begin{abstract}
The growing demand for intelligent processing at the edge of IoT networks is constrained by the severe computational and memory limitations of microcontroller units, which render impractical conventional deep learning approaches. We propose a neuromorphic-inspired classifier based on the Receptron model, a single-unit architecture capable of implementing non-linearly separable decision boundaries, without resorting to multi-layer networks. The model is designed for direct deployment on mid-range MCUs, while supporting continuous on-device adaptation. Experimental evaluation on basic dataset benchmarks yields cross-validated accuracies compatible with standard machine learning method baselines. These results position the Receptron as a viable and interpretable alternative for resource-constrained neuromorphic edge systems operating in dynamic, non-stationary environments.
\end{abstract}


\section{Introduction}
The rapid proliferation of interconnected smart devices, combined with the growing maturity of Artificial Intelligence (AI) techniques, has generated an increasing demand for embedding intelligent processing capabilities directly within physical, real-world systems \cite{gill2022ai}.

For this reason, a paradigm shift from cloud AI to edge AI is currently taking place: this is based on the provisioning of intelligence to the remote devices that are the backbone of the IoT \cite{singh2023edge}. The deployment of architectures originally designed for GPU-equipped datacenters at the edge of the IoT is hampered by the fact that this environment is currently dominated by microcontroller units (MCUs). MCUs are constrained by limited computational power, memory, data transmission rates and energy resources, making traditional machine learning models based on ANNs unsuitable for direct implementation \cite{sanchez-iborra2020tinyml-enabled}. To overcome these limitations, techniques such as model quantization, pruning and optimization \cite{kong2022edge} have been developed. While these techniques improve feasibility, they also introduce trade-offs, including reduced accuracy, increased inference latency, or higher resource consumption \cite{sanchez-iborra2020tinyml-enabled}.

To enable efficient inference at the edge, a wide range of hardware-oriented solutions has been explored, including analog and digital in-memory computing and, more recently, neuromorphic approaches based on spiking neural networks (SNNs) \cite{shooshtari2026review}. Nevertheless, the problem of enabling adaptation and training directly on the device has received considerably less attention. This is particularly critical for edge AI systems, which must often cope with sensor drift, recalibration needs and noisy, unpredictable operating environments \cite{ilic2026adaptive}. Such limitations are intimately bound to the conventional ANN trained through backpropagation, whose resource requirements remain fundamentally incompatible with on-device learning. This is due to their architectures based on linear elemental building blocks (the perceptrons \cite{rosenblatt1958perceptron}) organized in multilayers to classify nonlinear inputs \cite{lecun2015deep}. In a multilayer perceptron network, the information coming to the input units is recoded into an internal representation and the outputs are generated by the internal representation rather than by the original pattern. The expressiveness and generalization ability of neural networks are dependent on their size, which makes them inherently complex and extremely energy hungry.

Here we take a different angle of view and we argue that a crucial property that an elemental unit of an artificial neural network should exhibit, in analogy with biological neurons, is nonlinearity \cite{kandel2000principles}. This property critically unlocks the potential of the neuromorphic approach, substantially reducing the problems related to the use of massive networks of linear building blocks such as perceptrons.

In this paper, we aim to fully exploit this nonlinearity by introducing a novel formulation of the Receptron model \cite{martini2022receptron}, a generalization of the perceptron \cite{paroli2023solving}, characterized by input-dependent weight functions \cite{martini2022receptron,paroli2023solving}. We show that Receptron-based networks are suited for hardware-constrained systems (MCUs) and they are capable of on-device adaptation while retaining a high degree of interpretability.

\section{Mathematical Formulation}

\subsection{Receptron Formulation and Hard-Thresholded Decision Boundaries}
The classical Perceptron partitions the feature space by means of a linear hyperplane, mapping inputs to a binary output through a step function \cite{rosenblatt1958perceptron}. However, the linearity of its decision boundary severely limits expressive power on non-linearly separable data \cite{minsky1969perceptron}. Multi-layer extensions resolve this limitation at the cost of higher computational complexity and reduced interpretability, both of which are undesirable in safety-critical deployments.

To overcome these constraints, a novel single-unit model, hereafter referred to as the Receptron, has been proposed \cite{martini2022receptron,paroli2023solving}, capable of realising non-linearly separable functions within a single computational unit \cite{martini2022receptron,paroli2023solving}. This behaviour is derived from a generalised summation rule in which the synaptic weights are themselves functions of the input vector, as formalised in \cite{paroli2023solving}. The resulting activation function of the Receptron is defined as:
\begin{equation}
  S(\mathbf{x}) = H\!\left(\sum_{i} \tilde{w}_i(\mathbf{x})\, x_i\right)
  \label{eq:receptron_activation}
\end{equation}
where $H$ denotes the Heaviside step function, $\mathbf{x} = (x_1, \ldots, x_n) \in \mathbb{R}^n$ is the input vector and $\tilde{w}_i(\mathbf{x})$ is the input-dependent weight associated with the $i$-th component. The summation $\sum_{i} \tilde{w}_i(\mathbf{x})\, x_i$ constitutes a non-linear inner product in the original feature space, enabling the unit to implement decision boundaries of arbitrary curvature without explicit kernel mapping.

To model highly non-linear, non-convex and disjoint decision boundaries without projecting the input space into an infinite-dimensional feature space, the Receptron is reformulated as a combination of Gaussian receptive fields.

Let $\Omega = \{c_1, \ldots, c_K\} \subset \mathbb{R}^n$ be a set of $K$ prototype centers, each associated with a bandwidth parameter $\sigma_k > 0$. The activation profile of the $k$-th receptive field is modelled by an isotropic Gaussian function:
\begin{equation}
  G_k(\mathbf{x}) = \exp\!\left(-\frac{\|\mathbf{x} - \mathbf{c}_k\|^2}{2\sigma_k^2}\right)
  \label{eq:gaussian}
\end{equation}

The adoption of an isotropic Gaussian (i.e., a scalar bandwidth $\sigma_k$ and a diagonal precision matrix) represents a useful simplification: it reduces the number of free parameters per center while preserving a satisfactory approximation of the local data geometry, provided that the centers are allocated with sufficient mutual separation.

This representation is formally equivalent to the original non-linear formulation of the Receptron and can be derived via a local multi-dimensional Taylor series expansion centered on a set of prototype points.

To establish the equivalence with the summation rule of Eq.~(\ref{eq:receptron_activation}), the Gaussian function is expanded in a Taylor series around the origin. For a given center $c_k = 0$ (the general case follows by translation), one obtains:
\begin{equation}
  G(\mathbf{x}) = 1 - a_1(\sigma)\|\mathbf{x}\|^2 + a_2(\sigma)\|\mathbf{x}\|^4 - a_3(\sigma)\|\mathbf{x}\|^6 + \cdots
  \label{eq:taylor}
\end{equation}

where the scalar coefficients $a_m(\sigma) = \frac{1}{m!\,(2\sigma^2)^m}$ are uniquely determined by $\sigma$ and encode the width of the receptive field.

From this formulation we can rewrite it into a new expression to make each summand linear in the input components:
\begin{equation}
  G(\mathbf{x}) = \sum_i \left[\sum_{m \geq 0} (-1)^m a_m(\sigma)\|\mathbf{x}\|^{2(m-1)}\right] x_i^2
  \label{eq:linear_form}
\end{equation}
\begin{equation}
  \begin{aligned}
    G(\mathbf{x}) &= \sum_i \tilde{w}_i(\mathbf{x})\, x_i, \\
    \tilde{w}_i(\mathbf{x}) &:= \sum_{m \geq 0} (-1)^m a_m(\sigma)\|\mathbf{x}\|^{2(m-1)} x_i
  \end{aligned}
  \label{eq:weight_equiv}
\end{equation}

Equation~(\ref{eq:weight_equiv}) confirms the formal equivalence between the receptive-field representation and the original Receptron summation rule: the input-dependent weight $\tilde{w}(\mathbf{x})$ emerging from the Taylor expansion is identical in structure to the weight function in Eq.~(\ref{eq:receptron_activation}). Furthermore, since the Gaussian envelope decays exponentially with distance from $c_k$, the cross-talk between distinct centers is negligible whenever the centers are separated by a distance significantly larger than their respective bandwidths, a condition that is enforced by the allocation strategy described in Section~\ref{ssec:centre_all}.

To map the combined non-linear outputs to a strict binary output, the global activation is passed through a hard-thresholding operator (Heaviside step function) $H$. Given a set of $K$ expansion centers $c_k$, associated bandwidths $\sigma_k$ and a global bias $\theta$, the classification function of the Receptron is defined as:
\begin{equation}
  f(\mathbf{x}) = H\!\left(\sum_{k=1}^{K} G_k(\mathbf{x}) - \theta\right)
  \label{eq:classification}
\end{equation}
where the hard-thresholding operator is defined as:
\begin{equation}
  H(z) = \begin{cases} 1 & \text{if } z \geq 0 \\ 0 & \text{otherwise} \end{cases}
  \label{eq:heaviside}
\end{equation}

By absorbing the bias $\theta$ into the activation argument, the decision boundary is implicitly defined as the level set:
\begin{equation}
  \left\{\mathbf{x} \in \mathbb{R}^n : \sum_{k=1}^{K} G_k(\mathbf{x}) = \theta\right\}
  \label{eq:level_set}
\end{equation}
Unlike the standard Perceptron, whose boundary is restricted to a flat linear manifold, the combination of multiple Taylor-expanded Gaussian envelopes allows the Receptron to construct arbitrary, complex isosurfaces at the level $\theta$. This mathematical structure enables the separation of intricately clustered data distributions, providing the model with a rich non-linear hypothesis space while preserving a deterministic, binary decision output. In addition, the Gaussian activation profile constitutes a natural match for noise-corrupted measurements of stationary physical states (such as sensor readings processed by edge-computing units) by exploiting the well-known optimality of Gaussian functions in the presence of additive white noise.

\subsection{Centre Allocation and Temporal Adaptability}
\label{ssec:centre_all}
A central design challenge in the proposed framework is the optimized allocation of the prototype centers $c_k$. In the proposed implementation, this problem is addressed through a self-organizing, neuromorphically-inspired methodology. Specifically, the centers are initialized by means of a class-conditional variant of Kohonen's Self-Organising Map (SOM) \cite{kohonen1998visual}, in which competitive learning is restricted to prototypes belonging to the same class label. This constraint preserves semantic separation between classes while optimizing intra-class prototype coverage. At each training step $t$, the Best Matching Unit (BMU), defined as the center $c^*(t)$ minimizing the Euclidean distance to the current input $x(t)$, is updated according to a Hebbian-like rule with a monotonically decreasing learning rate schedule. The class-conditional competition emulates a Winner-Takes-All dynamic analogous to the lateral inhibition observed in primary somatosensory and visual cortex \cite{hubel1977ferrier}, whilst the synaptic update rule replicates Hebbian plasticity, thereby grounding the model in a biologically plausible learning cycle.

After centers initialization, a core strength of the Receptron architecture is its mathematical flexibility with respect to temporal variations in the data distribution. In dynamic environments where data streams exhibit non-stationary statistics or concept drift, fixed-topology models are inherently subject to catastrophic forgetting or structural underfitting.

Consequently, the Receptron can translate its geometrical shape online, optimizing the trade-off between the plasticity required to assimilate data-drift streams and the stability necessary to retain established decision boundaries, a property that renders it particularly well-suited for continual learning in non-stationary, resource-constrained environments.

\section{Model Architecture}
\subsection{Initialization and Training}
For each class $c$, training samples are projected onto the leading principal subspace of dimension $\min(K, D, N_c)$. The $K$ prototype positions are distributed uniformly along each principal axis and back-projected into the original feature space. This ensures that the initial centers span the principal geometric ``backbone'' of each class, avoiding the collapsed or degenerate initializations that can afflict random centers in classical SOMs, particularly when classes occupy elongated or curved manifolds.

Then, at each epoch, training samples are randomly permuted. For each sample $\mathbf{x}$ with class label $c$, the BMU is identified among the centers assigned to class $c$ and updated as:
\begin{equation}
  c_{\mathrm{BMU}} \leftarrow c_{\mathrm{BMU}} + \eta(t)\cdot(\mathbf{x} - c_{\mathrm{BMU}}),
  \label{eq:som_update}
\end{equation}
where $\eta(t) = \eta_0 \cdot \gamma^t$ and $\gamma \in (0, 1)$ is the decay factor. The exponential schedule mirrors homeostatic synaptic plasticity: early epochs prioritize broad exploration of the feature space, while later epochs consolidate the prototype positions. The absence of a topological neighborhood structure, relative to classical SOMs, reduces the computational footprint and simplifies hardware mapping, exploiting a better parallelization of the process.

Upon convergence of the SOM phase, the bandwidth $\sigma_k$ of each center is estimated as the mean distance between $c_k$ and the training samples of its class, scaled by a user-controllable multiplier $\mu \in \mathbb{R}^+$. Centers embedded in dense regions of the training distribution receive smaller bandwidths, while those in sparser regions receive larger ones.

\subsection{Inference Mode and Classification}
After training, the inference procedure controls the presence of the input in the over-threshold region of the Receptron unit. At first, we check the presence of activation inside the region described by the implicit function in Eq.~(\ref{eq:level_set}); in practice we can achieve a simple and fast inference method with $K$ inner products and $K$ threshold comparisons and no dense matrix multiplications. In the more general case, we can have a population of receptrons for each class, associating each activated center $c_k$ to a vote with a value of $1/\sigma_k$ and aggregating scores by class; finally, the classifier chooses the class with the best score with an argmax function. This arithmetic profile is well suited to fixed-point implementations or to sparse event-driven computation models of neuromorphic cores.

\section{Experimental Evaluation}
\subsection{Experimental Setup}
We evaluate the classification efficacy of the proposed classifier using two standard benchmarks obtained from the scikit-learn library: the Iris dataset (150 samples, 4 features, 3 classes) \cite{iris_53} and the Breast Cancer Wisconsin dataset (569 samples, 30 features, 2 classes) \cite{breast_cancer_wisconsin_diagnostic_17}. Prior to model training, all input features were standardized to exhibit zero mean and unit variance. To ensure statistical validation and robustness, a five-fold cross-validation scheme was executed across the fully standardized datasets utilizing the proposed architecture.

The optimal hyperparameter configurations were established specifically for each classification task to match the underlying data distribution. For the Iris dataset, the architecture was configured with $K = 4$ cluster centers per class, 300 SOM training epochs, an initial learning rate $\eta_0 = 0.4$, a decay coefficient $\gamma = 0.995$, a bandwidth multiplier $\mu = 0.75$ and a hard threshold $\tau = 1.5$. In contrast, the configuration for the Breast Cancer dataset utilized $K = 8$ cluster centers per class, 400 training epochs, an initial learning rate $\eta_0 = 0.3$, a decay coefficient $\gamma = 0.995$, a bandwidth multiplier $\mu = 0.5$, and a hard threshold $\tau = 2.0$.

\subsection{Classification Results}
The overall classification performance achieved via the weighted vote discrimination procedure is summarized in Table~\ref{tab:results}. On the Iris dataset we obtain a value of 90.0\% accuracy compared with a baseline accuracy of 94.7\% for Support Vector Classification and 89.4\% for Random Forest Classification \cite{iris_53}. On the Breast Cancer dataset we obtain a value of 93.5\% accuracy compared to 94.4\% for Support Vector Classification and 97.9\% for Random Forest Classification \cite{breast_cancer_wisconsin_diagnostic_17}.

\begin{table}[!t]
    \centering
    \caption{Test-set accuracy and 5-fold cross-validation scores}
    \label{tab:results}
    \begin{tabular}{lcc}
        \toprule
        Dataset & Centers / Class & 5-fold CV ($\mu \pm \sigma$) \\
        \midrule
        Iris (3 classes)          & 4 & $90.0 \pm 5.1\,\%$ \\
        Breast Cancer (2 classes) & 8 & $93.5 \pm 1.1\,\%$ \\
        \bottomrule
    \end{tabular}
\end{table}

Detailed class-specific test metrics, including precision, recall and F1-score, are reported in Table~\ref{tab:iris_metrics} for the Iris dataset and Table~\ref{tab:bc_metrics} for the Breast Cancer Wisconsin dataset.

\begin{table}[!t]
    \centering
    \caption{Class-specific test-set metrics for the Iris dataset}
    \label{tab:iris_metrics}
    \begin{tabular}{ccccc}
        \toprule
        Iris Class & Precision (\%) & Recall (\%) & F1-score (\%) & Support \\
        \midrule
        0 & 100 & 100 & 100 & 11 \\
        1 &  73 &  85 &  79 & 13 \\
        2 &  82 &  69 &  75 & 13 \\
        \bottomrule
    \end{tabular}
\end{table}

\begin{table}[!t]
    \centering
    \caption{Class-specific test-set metrics for the Breast Cancer Wisconsin dataset (0 = Malignant, 1 = Benign)}
    \label{tab:bc_metrics}
    \begin{tabular}{lcccc}
        \toprule
        BC Class & Precision (\%) & Recall (\%) & F1-score (\%) & Support \\
        \midrule
        0 & 90 & 87 & 88 & 53 \\
        1 & 92 & 94 & 93 & 90 \\
        \bottomrule
    \end{tabular}
\end{table}

\section{Sensitivity Analysis}
\begin{figure}[!t]
  \centering
  \includegraphics[width=\linewidth]{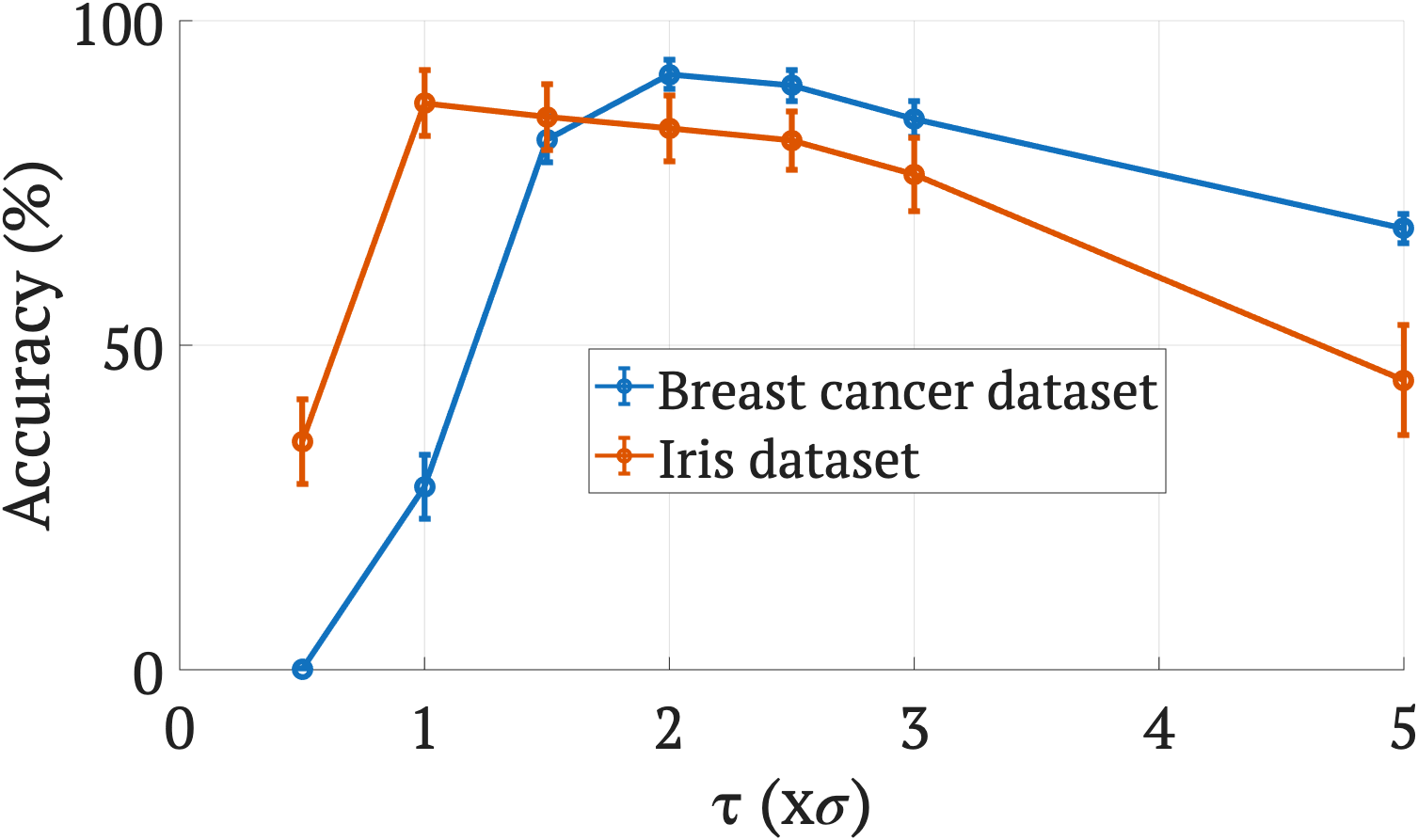}
  \caption{Average classification accuracy as a function of the threshold parameter $\tau$. Mean value and standard deviation on 100 repetitions with random separation of datasets (75\% training, 25\% testing).}
  \label{fig:acc_w_t}
\end{figure}

Figure~\ref{fig:acc_w_t} shows the classification accuracy as a systemic function of the boundary threshold parameter $\tau$ (normalized in units of $\sigma$).

When an overly restrictive threshold is applied ($\tau \le 1$), the activation radius of each cluster center encompasses only a minor fraction of the empirical intra-class dispersion. This configuration induces a high rate of unclassified test instances that fall entirely outside the receptive fields of all available centers. Minimizing the number of data points outside serves as a key optimization criterion; a zero or near-zero unclassified-sample count means that the model comprehensively encapsulates the relevant feature space. As $\tau$ scales upward, the classification accuracy exhibits a sharp monotonic increase before reaching a stable steady-state plateau. The optimal trade-off between categorization accuracy and feature space coverage is achieved at $\tau = 1.5$ for the Iris dataset and $\tau = 2.0$ for the Breast Cancer dataset. Conversely, in the asymptotic limit ($\tau \to \infty$, representing an unbounded, thresholdless classifier configuration), the accuracy degrades substantially. This degradation is driven by the spatial overlapping of competing decision regions, an effect that can only be mitigated by the weighted vote procedure.

\begin{figure}[!t]
  \centering
  \includegraphics[width=\linewidth]{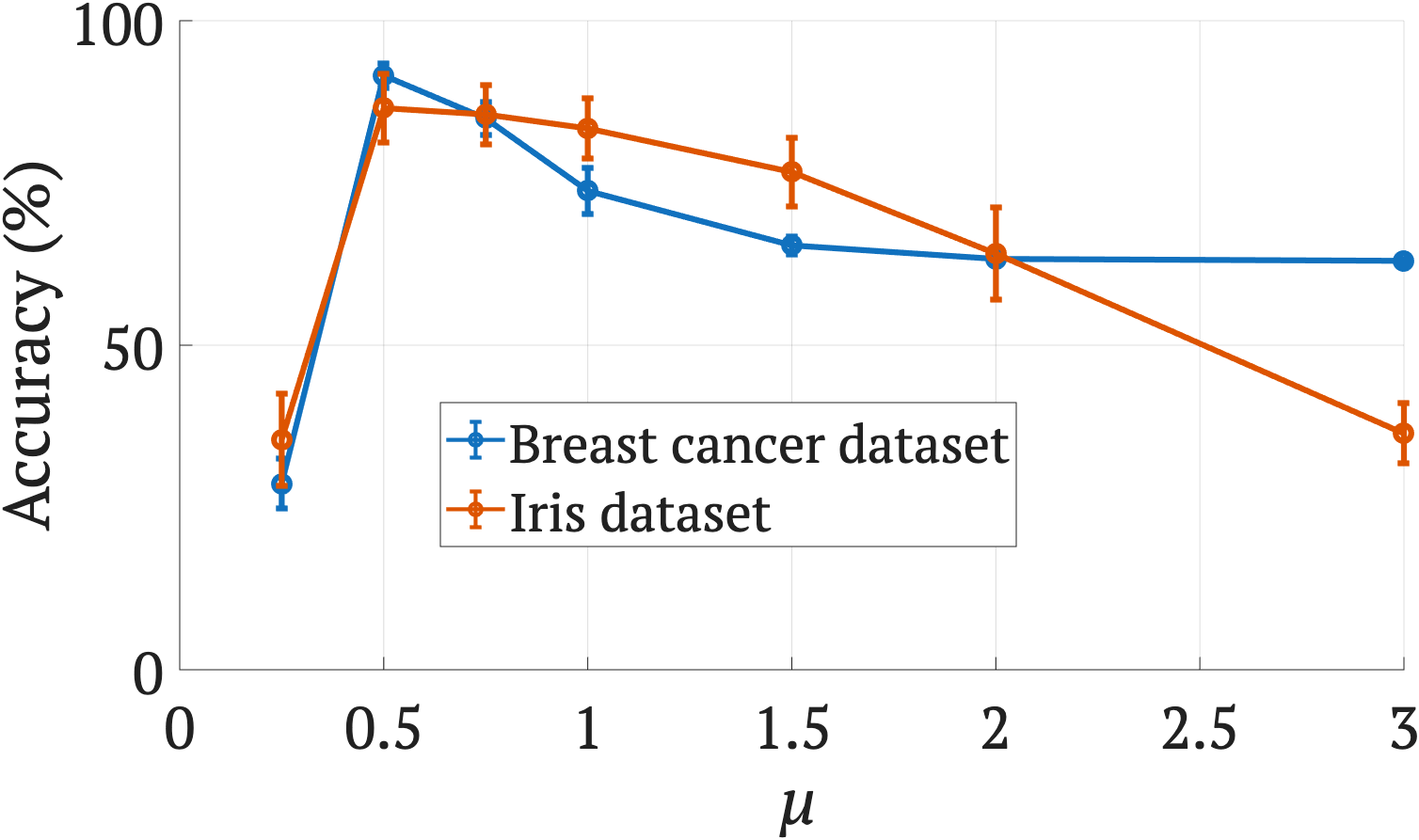}
  \caption{Average classification accuracy as a function of the bandwidth multiplier parameter $\mu$. Mean value and standard deviation on 100 repetitions with random separation of datasets (75\% training, 25\% testing).}
  \label{fig:acc_w_m}
\end{figure}

The model sensitivity with respect to the bandwidth multiplier $\mu$ governs the spatial conformity between the learned implicit separation and the empirical training distribution. This modulation serves to calibrate the algorithmic density estimation. The experimental results compiled in Figure~\ref{fig:acc_w_m} reveal a unimodal profile, with a performance peak centered near $\mu = 0.5$. This indicates that the uncalibrated analytical bandwidth calculation inherently tends to overestimate the true dispersion parameters within these specific spaces. For the Breast Cancer dataset, a tighter boundary constraint (lower $\mu$) relative to the Iris benchmark is selected; in clinical screening frameworks, for example, such strict spatial boundaries can favor rigid adherence to canonical pathological signatures.

\section{Discussion}
\subsection{Computational Footprint}
The inference cost of the proposed classifier is fundamentally governed by $K \cdot n$ multiply-accumulate (MAC) operations. For the Iris configuration ($K = 12$ total centres, $n = 4$ features), this translates to a deterministic computational overhead of 48 MAC and 12 comparisons per evaluation sample. For the Breast Cancer benchmark ($K = 16$, $n = 30$), the higher input dimensionality scales the arithmetic burden proportionally, yet the overall computational load remains well within the strict hardware constraints of low-power microcontrollers.

Furthermore, the weighted vote routine can be streamlined into a lightweight memory retrieval operation followed by a class-wise summation. This implementation route is highly compatible with low-complexity architectures. In optimal, non-overlapping topological scenarios, the classification boundary simplifies to a binary response for each class, enabling rapid and efficient discrimination.

\subsection{Parameter Optimization}
Optimizing the number of cluster centers $K$ per class represents a primary design challenge. In this work, suitable $K$ values were identified empirically through a rapid, low-iteration search starting from baseline heuristics. Future iterations of this architecture will incorporate adaptive algorithms to automate the discovery of optimal cluster counts based on local density metrics.

The sensitivity analysis highlights an overestimation of the analytical bandwidth ($\sigma$). This effect is expected given that the empirical cluster centers extracted from physical or complex datasets deviate from a strict Gaussian distribution. If the local data points adhered to an ideal normal distribution, the calculated $\sigma$ values would exhibit higher fidelity, yielding an optimal $\mu$ closer to 1.0.

Nevertheless, the statistical approximation provides a robust baseline for modeling data spread around the centers, with the multiplier $\mu$ acting as a necessary correction factor for non-normal distributions or initialization offsets. With the optimized bandwidth multiplier, the ideal threshold falls within the $1.5\sigma$ to $2.0\sigma$ window, a regime that statistically bounds roughly 95\% of the sample population in standard normal distributions. This suggests that a Gaussian function provides excellent approximation capabilities across diverse statistical point spreading, a hypothesis to be further validated against broader multi-domain benchmarks.

\subsection{Online Training}
While standard application scenarios often assume stationary data profiles, physical deployments frequently suffer from sensor drift, thermal fluctuations, or require periodic recalibration. Re-baselining a physical sensor typically shifts its baseline operating point while preserving the underlying variance of its characteristic response.

Under the proposed model, such systematic drifts can be accommodated by proportionally translating the coordinates of the learned cluster centers. Consequently, each Receptron can execute on-device retraining with minimal computational overhead, completely bypassing the need for intensive global backpropagation loops. This low-complexity update mechanism enables continuous, online edge learning within resource-constrained hardware environments.

\subsection{Hardware Mapping Feasibility and Mid-Range Microcontroller Compatibility}
The inference algorithm maps naturally onto commercial mid-range MCUs such as the ARM Cortex-M4 family or the ESP32, which operate at 80--240 MHz with 64--256 KB SRAM and up to 1 MB Flash.
In the most demanding configuration evaluated (Breast Cancer: 16 centers, 30 features), storing each coordinate as a standard 32-bit single-precision floating-point value requires a mere 1,920 bytes ($\sim$1.9 KB) of memory. Even when accounting for variance vectors and threshold constants, the comprehensive model footprint remains well below 4 KB of Flash and RAM, under 2\% of available SRAM on a typical 128 KB device, leaving the remaining memory blocks entirely free for real-time operating systems (RTOS), communication network stacks such as LoRaWAN or Wi-Fi and sensor data ingestion buffers. Requiring 480 MAC operations per inference cycle yields a theoretical latency in the order of tens of microseconds on hardware-FPU-equipped cores.
Unlike deep learning architectures that mandate the storage of massive computation graphs and error gradient tensors, which renders them impractical for standard MCU deployment due to RAM saturation, the online update rule of the proposed classifier relies exclusively on linear vector translations, allowing the system to continuously track and compensate for physical sensor drift without interrupting primary edge-monitoring operations.

\section{Conclusions}
We have presented a neuromorphic-inspired classification framework that harmonizes competitive Self-Organizing Map (SOM) learning, thresholded Gaussian activation functions and PCA-guided initialization into a highly interpretable, computationally efficient architecture. The classification capability has been tested on the Iris (90.0\% CV) and Breast Cancer Wisconsin (93.5\% CV) benchmarks and confirms that this approach yields classification accuracies competitive with classical machine learning methods while requiring only a low number of arithmetic operations per inference cycle.

Sensitivity analysis over the parameters $\tau$ and $\mu$ establishes a stable operational optimum within the windows of $\tau \in [1.5, 2.0]$ and $\mu \approx 0.5$, for the tested datasets. Beyond raw accuracy, the architecture provides vital features for real-world deployments: seamless integration into low-cost commercial mid-range microcontrollers and an online update rule that facilitates continuous adaptation without the need for exhaustive global retraining. These dual properties position the proposed classifier as a viable candidate for hardware-constrained neuromorphic edge systems operating in dynamic, evolving environments. Future work will focus on the development of adaptive mechanisms for autonomous center ($K$) selection based on local class density, alongside rigorous validation on industrial-grade fault-detection benchmarks. The architecture's low memory footprint, deterministic latency and online adaptability make it directly applicable to a broad class of resource-constrained scenarios, including embedded predictive maintenance, portable point-of-care diagnostics and drift-compensating distributed sensing on LoRaWAN nodes.

\section*{Acknowledgment}

\end{document}